\documentclass{article}
\usepackage{spconf} 
\usepackage[english]{babel}
\usepackage{ifpdf}
\usepackage{cite} 
\usepackage{url}
\usepackage{hyperref}

\ifpdf
	\usepackage[pdftex]{graphicx}
	\graphicspath{{./figures/}}
\else
	\usepackage[dvips]{graphicx}
	\graphicspath{./figures/}
\fi
\usepackage{color}
\usepackage{pgf, tikz, pgfplots}
\usetikzlibrary{shapes, arrows, automata}
\usepackage{caption} 
\usepackage{subcaption} 
\usepackage{amsmath}
\usepackage{amsfonts, amssymb, amsthm}
\usepackage{mathrsfs}
\usepackage{upgreek}
\usepackage{algorithm,algpseudocode}
\usepackage{enumerate}
\usepackage{multirow}
\usepackage{rotating}
\input{mysymbol.sty}

\newcommand{\QED}{\hfill\ensuremath{\blacksquare}}

\name{Alex Nowak$^{\dag, \S}$, Soledad Villar$^{\ddag,\ast,\S}$, Afonso S Bandeira$^{\ddag,\ast}$ and Joan Bruna$^{\ddag,\ast}$ \thanks{JB was partially supported by Samsung Electronics (Improving Deep Learning using Latent Structure), DOA W911NF-17-1-0438. ASB was partially supported by NSF DMS-1712730 and NSF DMS-1719545. SV was partially supported by the Simons Algorithms and Geometry (A\&G) Think Tank. \newline
$\S$ \ \small{Equal Contribution}}}

\address{\dag \ Sierra Team, INRIA and Ecole Normale Superieure, Paris \\
\ddag \ Courant Institute, New York University, New York \\
$\ast$ \ Center for Data Science, New York University, New York \\
}
\title{Revised Note on Learning Quadratic Assignment with Graph Neural Networks}
\begin{document}
\ninept

\maketitle

\begin{abstract} 

 Inverse problems correspond to a certain type of optimization problems formulated over appropriate input distributions. Recently, there has been a growing interest in understanding the computational hardness of these optimization problems, not only in the worst case, but in an average-complexity sense under this same input distribution.

In this revised note, we are interested in studying another aspect of hardness, related to the ability to learn how to solve a problem by simply observing a collection of previously solved instances. These `planted solutions' are used to supervise the training of an appropriate predictive model that parametrizes a broad class of algorithms, with 
the hope that the resulting model will provide good accuracy-complexity tradeoffs in the average sense.

We illustrate this setup on the Quadratic Assignment Problem, a fundamental problem in Network Science. We observe that data-driven models based on Graph Neural Networks offer intriguingly good performance, even in regimes where standard relaxation based techniques appear to suffer. 


\end{abstract} 

\parskip 4pt

\setlength{\belowdisplayskip}{3pt} \setlength{\belowdisplayshortskip}{0pt}
\setlength{\abovedisplayskip}{3pt} \setlength{\abovedisplayshortskip}{0pt}

\section{Introduction}



Many tasks, spanning from discrete geometry to statistics,
are defined in terms of \emph{computationally hard} optimization problems. 
Loosely speaking, computational hardness appears when the algorithms 
to compute the optimum solution scale poorly with the problem size, say faster than any polynomial.
For instance, in high-dimensional statistics we may be interested in the 
task of estimating a given object from noisy measurements 
under a certain generative model.
In that case, the notion of hardness contains both a statistical aspect, 
that asks above which signal-to-noise ratio the estimation is feasible, and a computational one, that restricts the estimation to be computed in polynomial time.
An active research area in Theoretical Computer Science and Statistics is to understand 
the interplay between those statistical and computational detection thresholds; see~\cite{abbe17} 
and references therein for an instance of this program in the community detection problem, or~\cite{wainwright08,jordan13,Berthet_Rigollet_SparsePCAcolt} for examples of statistical inference tradeoffs under computational constraints.

Instead of investigating a designed algorithm for the problem in question, we consider a data-driven approach to learn algorithms from solved instances of the problem.
In other words, given a collection  
$(x_i, y_i)_{i \leq L}$ of problem instances drawn from a certain distribution, we ask whether one can \emph{learn} an algorithm that achieves good accuracy at solving new instances of the same problem -- also being drawn from the same distribution, and to what extent the resulting algorithm can reach those statistical/computational thresholds.

The general approach is to cast an ensemble of algorithms as  
neural networks $\hat{y}=\Phi(x;\theta)$ with specific architectures 
that encode prior knowledge on the algorithmic class, parameterized by
 $\theta \in \mathbb{R}^S$. The network is trained to minimize the empirical loss $\mathcal{L}(\theta) = L^{-1} \sum_{i} \ell(y_i, \Phi(x_i; \theta))$, for a given measure of error $\ell$, using 
stochastic gradient descent. This leads to yet another notion of \emph{learnability hardness}, that measures to what extent the problem can be solved with no prior knowledge of the specific 
algorithm to solve it, but only a vague idea of which operations it should involve. 

In this revised version of ~\cite{nowakICML} we focus on a particular NP-hard problem, the Quadratic Assignment Problem (QAP), and study data-driven approximations to solve it. Since the problem is naturally formulated in terms of graphs, we consider the so-called Graph Neural Network (GNN) model \cite{scarselli09}. This neural network alternates between applying linear combinations of local graph operators -- such as the graph adjacency or the graph Laplacian, and pointwise non-linearities, and has the ability to model some forms of non-linear message passing and spectral analysis, as illustrated for instance in the data-driven Community Detection methods in the Stochastic Block Model~\cite{bruna2017community}. Existing tractable algorithms for the QAP include spectral alignment methods~\cite{umeyama88} and methods based on semidefinite programming relaxations~\cite{zhao98, ferreira17}. 
Our preliminary experiments suggest that the GNN approach taken here may be able to outperform the spectral and SDP counterparts on certain random graph models, at a lower computational budget. 
We also provide an initial analysis of the learnability hardness by studying the optimization landscape of a simplified GNN architecture. Our setup reveals that, for the QAP, the landscape complexity is controlled  by the same concentration of measure phenomena that controls the statistical hardness; see Section \ref{landscapesec}.

The rest of the paper is structured as follows. Section~\ref{qapsect} presents the problem set-up and describes existing relaxations of the QAP. Section~\ref{gnnsect} describes the graph neural network architecture, Section \ref{landscapesec} presents our landscape analysis, and Section~\ref{expsections} presents our numerical experiments. Finally, Section~\ref{opensect} describes some open research directions motivated by our initial findings.


\section{Quadratic Assignment Problem}
\label{qapsect}
Quadratic assignment is a classical problem in combinatorial optimization. For $A,B$ $n\times n$ symmetric matrices it can be expressed as 
\begin{eqnarray} \operatorname{maximize}\, \operatorname{trace}(AXBX^\top), \label{qap}
\;\;
\text{subject to } X\in \Pi,
\end{eqnarray}
where $\Pi$ is the set of permutation matrices of size $n\times n$. 
Many combinatorial optimization problems can be formulated as quadratic assignment. For instance, the network alignment problem consists on given $A$ and $B$ the adjacency graph of two networks, to find the best matching between them, i.e.:
\begin{eqnarray}
\operatorname{minimize}\,  \|AX - XB\|_{F}^2, \label{qap2} \;\;
\text{subject to }  X \in \Pi.
\end{eqnarray}
By expanding the square in~\eqref{qap2} one can obtain an equivalent optimization of the form~\eqref{qap}. The value of~\eqref{qap2} is 0 if and only if the graphs $A$ and $B$ are isomorphic.
The minimum bisection problem can also be formulated as a QAP. This problem asks, given a graph $B$, to partition it in two equal sized subsets such that the number of edges across partitions is minimized. This problem, which is natural to consider in community detection, can be expressed as finding the best matching between $A$, a graph with two equal sized disconnected cliques, and $B$. 

The quadratic assignment problem is known to be NP-hard and also hard to approximate~\cite{pardalos94}. Several algorithms and heuristics had been proposed to address the QAP with different level of success depending on properties of $A$ and $B$~\cite{fiori2015spectral, lyzinski2016graph,onaran2017projected}. 
We refer the reader to~\cite{feizi16} for a recent review of different methods and numerical comparison. According to the experiments performed in~\cite{feizi16} the most accurate algorithm for recovering the best alignment between two networks in the distributions of problem instances considered below is a semidefinite programming relaxation (SDP) first proposed in~\cite{zhao98}. However, such relaxation requires to ‘lift’ the variable $X$ to an $n^2\times n^2$ matrix and solve an SDP that becomes practically intractable for $n>20$. Recent work~\cite{ferreira17} has further relaxed the semidefinite formulation to reduce the complexity by a factor of $n$, and proposed an augmented lagrangian alternative to the SDP which is significantly faster but not as accurate, and it consists of an optimization algorithm with $O(n^3)$ variables. 

There are known examples where the SDP is not able to prove that two non-isomorphic graphs are actually not isomorphic (i.e. the SDP produces pseudo-solutions that achieve the same objective value as an isomorphism but that do not correspond to permutations~\cite{o14, villar2016polynomial}).
Such adverse example consists on highly regular graphs whose spectrum have repeated eigenvalues, so-called \emph{unfriendly graphs}~\cite{aflalo2014graph}.
We find QAP to be a good case study for our investigations for two reasons. It is a problem that is known to be NP-hard but for which there are natural statistical models of inputs, such as models where one of the graphs is a relabelled small random perturbation of the other, on which the problem is believed to be tractable. On the other hand, producing algorithms capable of achieving this task for large perturbations appears to be difficult.  It is worth noting that, for statistical models of this sort, when seen as inverse problems, the regimes on which the problem of recovering the original labeling is possible, impossible, or possible but potentially computationally hard are not fully understood.

 \vspace{2cm}
\section{Graph Neural Networks}
\label{gnnsect}

The Graph Neural Network, introduced in \cite{scarselli09} and further simplified in \cite{li15, duvenaud2015convolutional, sukhbaatar16} is a neural network architecture based 
on local operators of a graph $G = (V, E)$, offering a powerful balance 
between expressivity and sample complexity; see \cite{DBLP:journals/corr/BronsteinBLSV16} for a recent survey on models and applications of deep learning on graphs.

Given an input signal $F \in \mathbb{R}^{V \times d}$ on the vertices of $G$, we consider graph intrinsic linear operators that act locally on this signal:
The \textit{adjacency operator} is the map $A: F \mapsto A(F)$ where 
$(A F)_i:= \sum_{j\sim i }F_j~, $ with $i \sim j$ iff $(i,j) \in E$. The \textit{degree operator} is the diagonal linear map $D(F) = \text{diag}( A {\bf 1}) F~.$ Similarly, $2^J$-th powers of $A$, $A_J = \min(1, A^{2^J})$
encode $2^J$-hop neighborhoods of each node, and allow us to aggregate local information at different scales, which is useful in regular graphs.
We also include the 
average operator $ (U(F))_i = \frac{1}{|V|} \sum_j F_j$, which allows to broadcast information globally at each layer, thus giving the GNN the ability to recover average degrees, or more generally moments of local graph properties.
By denoting $\mathcal{A} = \{ {\bf 1}, D, A,A_1 \ldots, A_J, U \}$ the \emph{generator} family,   
 a GNN layer receives as input a signal $x^{(k)} \in \mathbb{R}^{V \times d_k}$ and produces $x^{(k+1)} \in \mathbb{R}^{V \times d_{k+1}}$ as \\ 
{\small 
\begin{equation}
\label{gnneq}
x^{(k+1)} = \rho \left( \sum_{B \in \mathcal{A}} B x^{(k)} \theta_{B}^{(k)} \right)~,
\end{equation}}
where $\Theta=\{ \theta_1^{(k)}, \dots, \theta_{|\mathcal{A}|}^{(k)} \}_k$, 
${\theta}_B^{(k)} \in \mathbb{R}^{d_k \times d_{k+1}}$, are trainable parameters and $\rho(\cdot)$ is a point-wise non-linearity, chosen in this work to be $\rho(z) = \max(0,z)$ for the first $d_{k+1}/2$ output coordinates and $\rho(z) =z$ for the rest.
We consider thus a layer with concatenated ``residual connections" \cite{he2016deep}, to both ease with the optimization when using large number of layers, but also to give the model the ability to perform  power iterations.
Since the spectral radius of the learned linear operators in (\ref{gnneq}) can grow as the optimization progresses, the cascade of GNN layers can become unstable to training. In order to mitigate this effect, we use spatial batch normalization \cite{ioffe2015batch} at each layer. 
The network depth is chosen to be of the order of the graph diameter, so that all nodes obtain information from the entire graph. In sparse graphs with small diameter, this architecture offers excellent scalability and computational complexity.


Cascading layers of the form (\ref{gnneq}) gives us the ability to approximate a broad family of graph inference algorithms, including some forms of spectral estimation. Indeed, power iterations are recovered by bypassing the nonlinear components and sharing the parameters across the layers. Some authors have observed \cite{gilmer2017neural} that GNNs are akin to message passing algorithms, although the formal connection has not been established. GNNs also offer natural generalizations to process high-order interactions, for instance using graph hierarchies such as line graphs \cite{bruna2017community} or using tensorized representations of the permutation group \cite{kondor2018covariant}, but these are out of the scope of this note.


The choice of graph generators encodes prior information on the nature of the estimation task. 
For instance, in the community detection task, the choice of generators is motivated by a model from Statistical Physics, the Bethe free energy \cite{saade2014spectral, bruna2017community}. In the QAP, one needs generators that are able to detect distinctive and stable local structures. Multiplicity of the spectrum of the graph adjacency operator is related to the (un)effectiveness of certain relaxations \cite{aflalo2014graph, lyzinski2016graph} (the so-called \emph{(un)friendly graphs}), suggesting that generator families $\mathcal{A}$ that contain non-commutative operators may be more robust on such examples.



\section{Landscape of Optimization}
\label{landscapesec}

\vspace{-0.2cm}

In this section we sketch a research direction to study the landscape of the optimization problem by studying a simplified setup. Let us assume that $A$ is the $n\times n$ adjacency matrix of a random weighted graph (symmetric) from some distribution $\mathcal D_n$, and let $B=\Pi A\Pi^{-1} + \nu$ where $\Pi$ is a permutation matrix and $\nu$ represents a symmetric noise matrix.

For simplicity let's assume that the optimization problem consists in finding a polynomial operator of degree $d$. If $A \in \mathbb R^{n\times n}$ denotes the adjacency matrix of a graph, then its embedding $E_A\in\mathbb R^{n\times k}$ is defined as 
\begin{equation*}E_A=P_{\beta}(A)Y \; \text{ where } P_{\beta^{(t)}}(A)= \sum_{j=0}^d \beta^{(t)}_j A^j, \; t=1,\ldots, k,\end{equation*}
and 
$P_{\beta}(A)Y = (P_{\beta^{(1)}}(A)y_1,\ldots, P_{\beta^{(k)}}(A)y_k)\in \mathbb R^{n\times k}$
such that $Y$ is a random $n\times k$ matrix with independent Gaussian entries with variance $\sigma^2$ and $y_t$ is the $t$-th column of $Y$. Each column of $Y$ is thought as a random initialization vector for an iteration resembling a power method. This model is a simplified version of \eqref{gnneq} where $\rho$ is the identity and $\mathcal{A} = \{ A \}$. 
Note that $P_\beta(\Pi A \Pi^{-1}) = \Pi P_\beta(A) \Pi^{-1}$ so it suffices to analyze the case where $\Pi=I$. 

We consider the following loss 
$$L(\beta)= - \mathbb E_{A,B,Y}\frac{\langle P_\beta(A)Y, P_\beta(B)Y \rangle}{ \|P_\beta(A)Y\|^2 + \|P_\beta(B)Y\|^2}.$$
Since $A,B$ are symmetric we can consider $e_1,\ldots,e_n$ an eigenbasis for $A$ with respective eigenvalues $\lambda_1,\ldots,\lambda_n$ and an eigenbasis for $B$, $\tilde e_1,\ldots,\tilde e_n$ with eigenvalues $\tilde\lambda_1,\ldots,\tilde\lambda_n$. Decompose $y^{(t)}=\sum_{i}y_i^{(t)} e_i = \sum_{i'} \tilde y_{i'}^{(t)} \tilde e_{i'}$. Observe that 
$$\langle P_\beta(A)Y, P_\beta(B)Y \rangle =\sum_{t=1}^k {\beta^{(t)}}^\top Q(A,B)^{(t)} \beta^{(t)}$$
where 
\begin{eqnarray*}Q(A,B)^{(t)}_{rs}&=& \langle A^r y^{(t)}, B^s y^{(t)}  \rangle \\ &=& \left(\sum_{i=0}^n\lambda_{i}^r y_i^{(t)}e_i\right)^\top \left(\sum_{i'=0}^n \tilde \lambda_{i'}^s \tilde y_{i'}^{(t)}\tilde e_{i'}\right) \\
&=& \sum_{i,i'} \lambda_i^r \tilde \lambda_{i'}^s y_i^{(t)} \tilde y_{i'}^{(t)} \langle e_i, \tilde e_{i'} \rangle.
\end{eqnarray*}
Note we can also consider a symmetric version of $Q$ that leads to the same quadratic form.
We observe that 
\begin{equation} \label{eqloss}
L(\beta) = \displaystyle - \mathbb E_{A,B,Y} \frac{\beta^\top Q(A,B) \beta}{\beta^\top (Q(A,A)+Q(B,B)) \beta}.
\end{equation}

For fixed $A,B$ (let's not consider the expected value for a moment), the quotient of two quadratic forms $\frac{\beta^\top R \beta}{\beta^\top S \beta}$ (where $S$ is a positive definite with Cholesky decomposition $S=CC^\top$) is maximized by the top eigenvector of ${C^{-1}RC^\top}^{-1}$. 

Say that $A$ is appropriately normalized (for instance let's say $A=n^{-3/2} W$ with $W$ a Wigner matrix, i.e. a random symmetric matrix with i.i.d. normalized gaussian entries). When $n$ tends to infinity the denominator of \eqref{eqloss} rapidly concentrates around its expected value
$$\mathbb E_Y Q(A,A)_{rs}=\sum_i \lambda_i^{r+s}\sigma^2 \to \frac{\sigma^2}{2\pi} \int_{-2}^2 \lambda^{r+s} (4-|\lambda|^2)_{+}^{1/2}d\lambda,$$
where the convergence is almost surely according to the semicircle law (see for instance \cite{tao}).

If $A\sim \mathcal D_n$ for large enough $n$, due to concentration of measure, with high probability we have
$$\beta^\top \mathbb E_Y Q(A,A) \beta(1-\epsilon) \leq \beta^\top  Q(A,A) \beta \leq \beta^\top \mathbb E_Y Q(A,A) \beta (1+\epsilon),$$
and similarly for $Q(B,B)$ (assuming that the noise level in $B$ is small enough to allow concentration). Intuitively we expect that the denominator concentrates faster than the numerator due to the latter's dependency on the inner products $\langle e_i, \tilde e_{i'} \rangle$. A formal argument is needed to make this statement precise. 
With this in mind we have 
\begin{multline*}
-(1-\epsilon)^{-1}\frac{ \beta^\top \mathbb \{E_{A,B,Y} Q(A,B)\} \beta}{\beta^\top ( \mathbb E_Y Q(A,A)+\mathbb E_Y Q(B,B) )\beta} \leq  L(\beta) \\ \leq -(1+\epsilon)^{-1}\frac{ \beta^\top \{\mathbb E_{A,B,Y}Q(A,B)\} \beta}{\beta^\top ( \mathbb E_Y Q(A,A)+\mathbb E_Y Q(B,B) )\beta}.\end{multline*}
If $\epsilon=0$, since the denominator is fixed one can use the Cholesky decomposition of $S=\mathbb E_Y Q(A,A)+\mathbb E_Y Q(B,B)$ to find the solution, similarly to the case where $A$ and $B$ are fixed. In this case, the critical points are in fact the eigenvectors of $Q(A,B)$, and 
one can show that $L(\beta)$ has no poor local minima, since it amounts to optimizing a quadratic form on the sphere. When $\epsilon>0$, one could expect that the concentration of the denominator around its mean somewhat controls the presence of poor local minima of $L(\beta)$. An initial approach is to use this concentration to bound the distance between the critical points of $L(\beta)$ and those of the `mean field' equivalent
$$g=-  \frac{\beta^\top \mathbb E_{A,B,Y} Q(A,B) \beta}{\beta^\top \mathbb E_{A,B,Y} (Q(A,A)+Q(B,B)) \beta}~.$$


If we denote $\tilde{S}=Q(A,A)+Q(B,B)$, we have
\begin{multline*}
\|\nabla_\beta L(\beta) - \nabla_\beta g\| \\
\leq \frac{1}{2}\left|(\beta^T\tilde{S}\beta)^{-1}
-(\beta^TS\beta)^{-1}\right|\|
\mathbb{E}_{A,B,Y}Q(A,B)\beta\| + \\
\frac{1}{2}|L(\beta)|\left\|(1+\epsilon)
\frac{\tilde{S}\beta}{\beta^T\tilde{S}\beta}
- \frac{S\beta}{\beta^TS\beta}\right\|
\end{multline*}
A more rigorous analysis shows that both terms in the upper bound tend to zero as $n$ grows due to the fast concentration of $\tilde{S}$ around its mean $S$. Another possibility is to rely instead on topological tools such as those developed in \cite{venturi2018neural} to control the presence of energy barriers as a function of $\epsilon$. 

\vspace{-0.4cm}
\section{Numerical Experiments} \label{expsections}
\begin{figure*}
    \centering
    \includegraphics[width=0.35\textwidth]{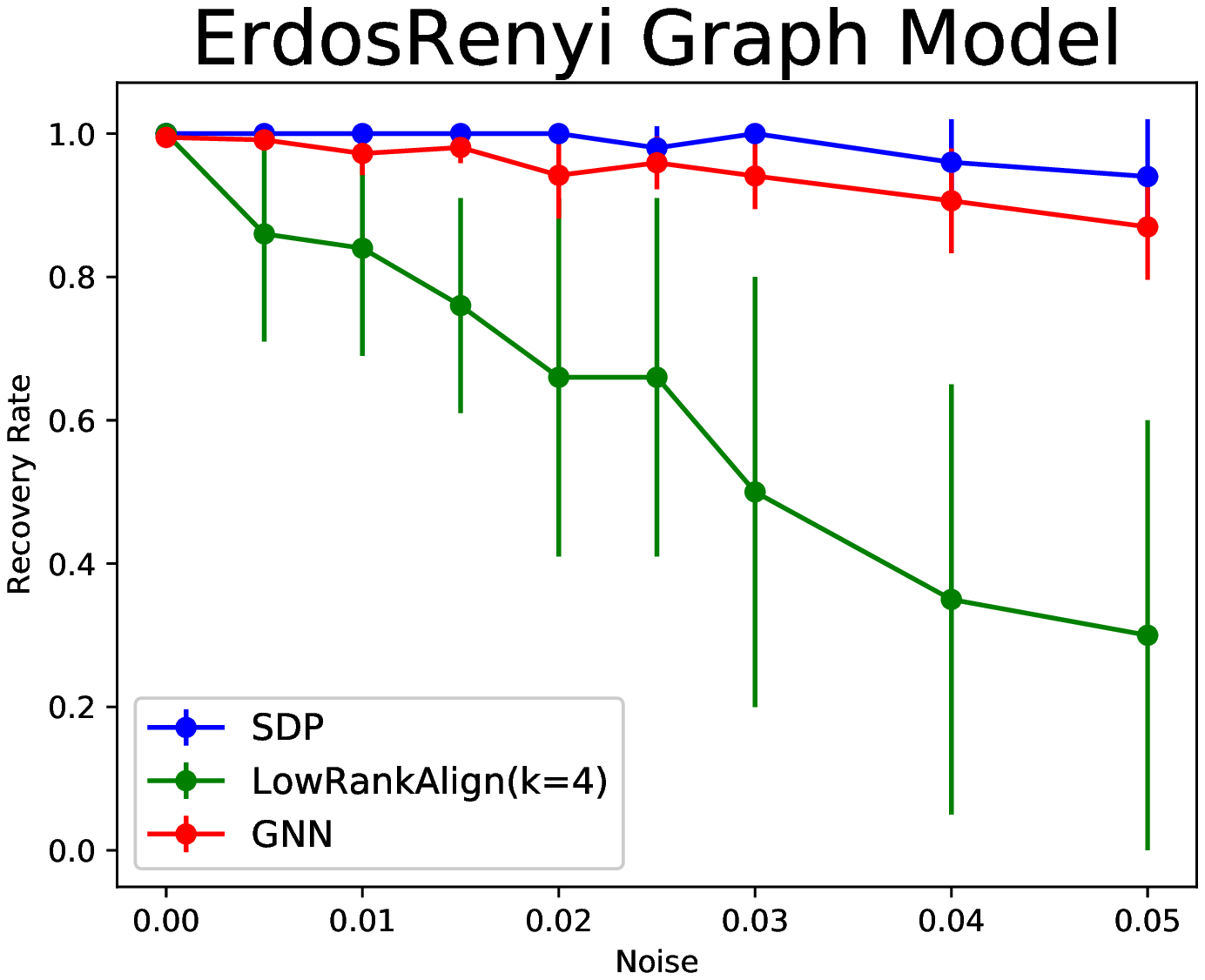}
	\includegraphics[width=0.35\textwidth]{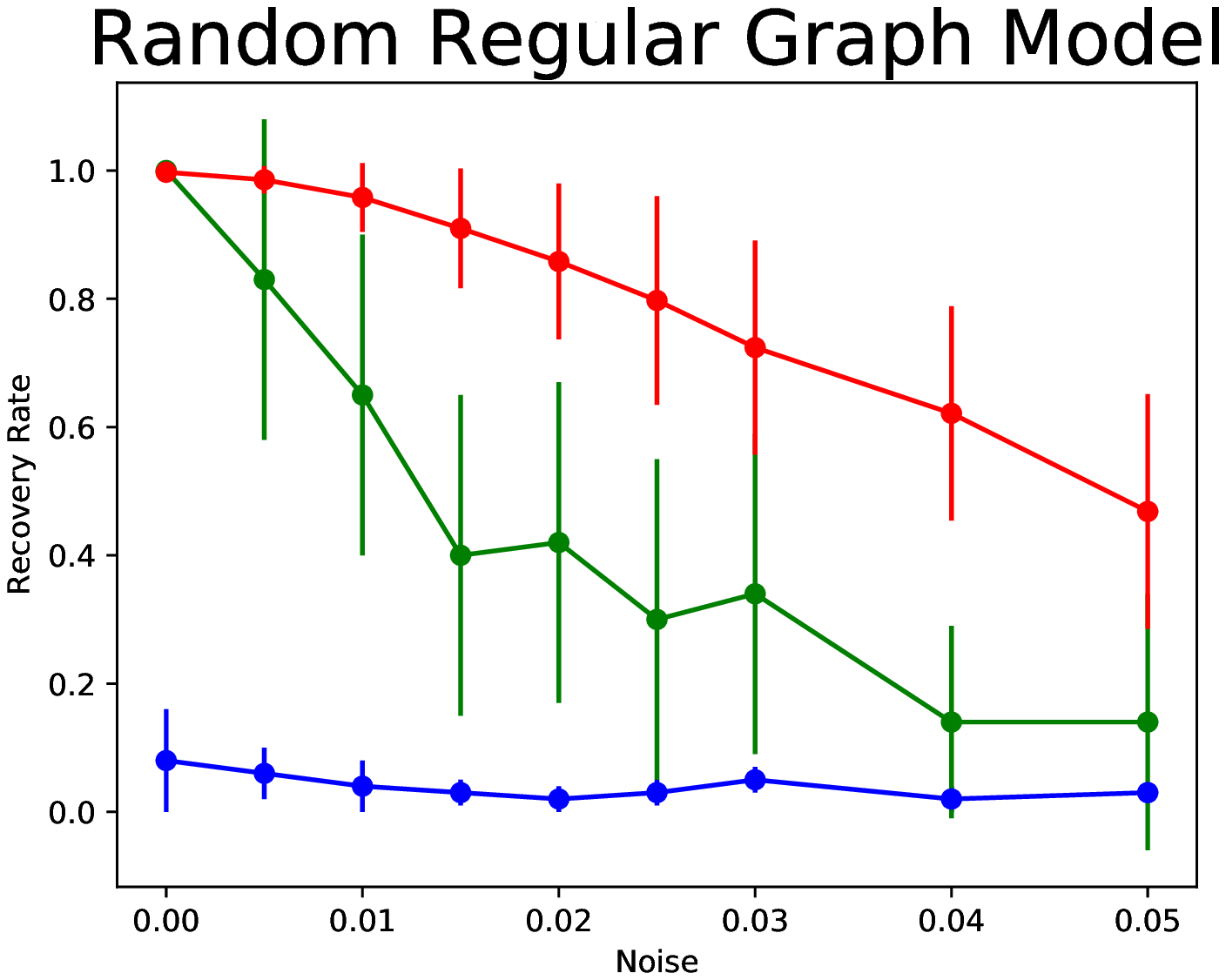}
    \caption{\label{fig.QAP} Comparison of recovery rates for the SDP \cite{peng10}, LowRankAlign \cite{feizi16} and our data-driven GNN, for the Erdos-Renyi model (left) and random regular graphs (right). All graphs have $n=50$ nodes and edge density $p=0.2$. The recovery rate is measured as the average number of matched nodes from the ground truth. Experiments have been repeated 100 times for every noise level except the SDP, which have been repeated 5 times due to its high computational complexity.}
\end{figure*}

We consider the GNN and train it to solve random planted problem instances of the QAP. Given a pair of graphs $G_1,G_2$ with $n$ nodes each, 
we consider a \emph{siamese} GNN encoder producing normalized embeddings 
$E_1,E_2\in\mathbb{R}^{n\times d}$. 
Those embeddings are used to predict 
a matching as follows. We first compute the
outer product $E_1 E_2^T$, that we then map 
to a stochastic matrix by taking the softmax along each row/column.  
Finally, we use standard cross-entropy loss to predict the corresponding
permutation index.
We perform experiments of the proposed data-driven model for the graph matching problem  \footnote{\scriptsize{Code available at \url{https://github.com/alexnowakvila/QAP_pt}}}. Models are trained using Adamax \cite{kingma2014adam} with $lr=10^{-3}$ and batches of size 32. We note that the complexity of this algorithm is at most $O(n^2)$.

\subsection{Matching Erdos-Renyi Graphs} \label{ER}
In this experiment, we consider $G_1$ to be a random Erdos-Renyi graph with edge density $p_e$. The graph $G_2$ is a small perturbation of $G_1$ according to the following error model considered in \cite{feizi16}:
\begin{equation} \label{noisemodel} 
G_2= G_1\odot(1-Q) +(1-G_1)\odot Q'
\end{equation}
where $Q$ and $Q'$ are binary random matrices whose entries are drawn from i.i.d. Bernoulli distributions such that $\mathbb P(Q_{ij}=1)=p_e$ and $\mathbb P(Q'_{ij}=1)=p_{e_2}$ with $p_{e_2}=p_e\frac{p}{p-1}$.
The choice of $p_{e_2}$ guarantees that the expected degrees of $G_1$ and $G_2$ are the same. We train a GNN with 20 layers and 20 feature maps per layer on a data set of 20k examples. We fix the input embeddings to be the degree of the corresponding node. In Figure~\ref{fig.QAP} we report its performance in comparison with the SDP from \cite{peng10} and the LowRankAlign method from \cite{feizi16}.


\subsection{Matching Random Regular Graphs}
Regular graphs are an interesting example because they tend to be considered harder to align due to their more symmetric structure. Following the same experimental setup as in \cite{feizi16}, $G_1$ is a random regular graph generated using the method from \cite{kim2003} and $G_2$ is a perturbation of $G_1$ according to the noise model (\ref{noisemodel}). Although $G_2$ is in general not a regular graph, the ``signal" to be matched to, $G_1$, is a regular graph. Figure \ref{fig.QAP} shows that in that case, the GNN is able to extract stable and distinctive features, outperforming the non-trainable alternatives. We used the same architecture as \ref{ER}, but now, due to the constant node degree, the embeddings are initialized with the 2-hop degree.

\section{Discussion}
\label{opensect}

Problems are often labeled to be as hard as their hardest instance. However, many hard problems can be efficiently solved for a large subset of inputs. 
This note attempts to learn an algorithm for QAP from solved problem instances drawn from a distribution of inputs. The algorithm's effectiveness is evaluated by investigating how well it works on new inputs from the same distribution.
This can be seen as a general approach and not restricted to QAP. In fact, another notable example is the community detection problem under the Stochastic Block Model (SBM)~\cite{bruna2017community}.
That problem is another particularly good case of study because there exists very precise predictions for the regimes where the recovery problem is (i) impossible, (ii) possible and efficiently solvable, or (iii) believed that even though possible, may not be solvable in polynomial time.

If one believes that a problem is computationally hard for most instances in a certain regime, then this would mean that no choice of parameters for the GNN could give a good algorithm. However, even when there exist efficient algorithms to solve the problem, it does not mean necessarily that an algorithm will exist that is expressible by a GNN. On top of all of this, even if such an algorithm exists, it is not clear whether it can be learned with Stochastic Gradient Descent on a loss function that simply compares with known solved instances. However, experiments in~\cite{bruna2017community} suggest that GNNs are capable of learning algorithms for community detection under the SBM essentially up to optimal thresholds, when the number of communities is small. 
Experiments for larger number of communities show that GNN models are currently unable to outperform Belief-Propagation, a baseline tractable estimation that achieves optimal detection up to the computational threshold. Whereas this preliminary result is inconclusive, it may guide future research attempting to elucidate whether the limiting factor is indeed a gap between statistical and computational threshold, or between learnable and computational thresholds.

The performance of these algorithms depends on which operators are used in the GNN. Adjacency matrices and Laplacians are natural choices for the types of problem we considered, but different problems may require different sets of operators. A natural question is to find a principled way of choosing the operators, possibly querying graph hierarchies.
Going back to QAP, it would be interesting to understand the limits of this problem, both statistically \cite{onaran2016}, but also computationally. In particular the authors would like to better understand the limits of the GNN approach and more generally of any approach that first embeds the graphs, and then does linear assignment. 


In general, understanding whether the regimes for which GNNs produce meaningful algorithms matches believed computational thresholds for some classes of problems is, in our opinion, a thrilling research direction.
It is worth noting that this approach has the advantage that the algorithms are learned automatically. However, they may not generalize in the sense that if the GNN is trained with examples below a certain input size, it is not clear that it will be able to interpret much larger inputs, that may need larger networks. This question requires non-asymptotic  deviation bounds, which are for example well-understood on problems such as the `planted clique'. 




\bibliography{example_paper}
\bibliographystyle{plain}

\end{document}